%% file: main.tex
\documentclass[runningheads]{llncs}
\usepackage[letterpaper, left=1in, right=1in, top=1in, bottom=1in]{geometry} 
 
\usepackage[year=2026]{eccv}
\usepackage[numbers]{natbib}
\usepackage{multirow}
\usepackage{tabularx}
\usepackage{wrapfig}

\usepackage{capt-of}
\usepackage{tcolorbox}
\tcbuselibrary{skins, breakable, listings}
\usepackage{listings}
\usepackage{enumitem}

\definecolor{headerBg}{HTML}{B0A878}
\definecolor{cardBg}{HTML}{FDFAF0}
\definecolor{cardFrame}{HTML}{D6CFA8}
\definecolor{codeFrame}{HTML}{C0C0C0}

\newtcolorbox{promptcard}[1]{%
  enhanced,
  width        = \linewidth,
  colback      = cardBg,
  colframe     = cardFrame,
  boxrule      = 0.6pt,
  arc          = 2pt,
  left         = 10pt,
  right        = 10pt,
  top          = 4pt,
  bottom       = 6pt,
  fontupper    = \rmfamily\small,
  title        = {\bfseries\normalsize #1},
  coltitle     = white,
  colbacktitle = headerBg,
  toptitle     = 4pt,
  bottomtitle  = 4pt,
  before skip  = 6pt,
  after skip   = 6pt,
  attach boxed title to top left = {xshift=0pt, yshift=0pt},
  boxed title style = {%
    sharp corners, colframe=headerBg, boxrule=0pt, arc=2pt},
  before upper = {\setlength{\parskip}{4pt}%
                  \setlength{\parindent}{0pt}},
}

\lstdefinestyle{promptlst}{%
  basicstyle       = \ttfamily\scriptsize,
  breaklines       = true,
  breakatwhitespace= true,
  frame            = single,
  rulecolor        = \color{codeFrame},
  xleftmargin      = 0pt,
  framexleftmargin = 0pt,
  aboveskip        = 6pt,
  belowskip        = 6pt,
  columns          = fullflexible,
}

\usepackage{eccvabbrv}
\usepackage{dsfont}
\usepackage{graphicx}
\usepackage{booktabs}

\usepackage[accsupp]{axessibility}  
\usepackage{xspace}
\newcommand{\ours}{{MM-Zero}\xspace}

\usepackage{hyperref}

\usepackage{orcidlink}

\begin{document}

\title{\ours: Self-Evolving Multi-Model Vision Language Models From Zero Data} 

\titlerunning{MM-Zero}

\author{Zongxia Li\inst{1}\thanks{Corresponding authors: \email{zli12321@umd.edu}, \email{hongyang\_du@brown.edu}, \email{chengsong@wustl.edu}}, Hongyang Du\inst{2}$^\star$, Chengsong Huang\inst{3}$^\star$, Xiyang Wu\inst{1}\\ Lantao Yu\inst{4} Yicheng He\inst{5}, Jing Xie\inst{1}, Xiaomin Wu\inst{1}\\ Zhichao Liu\inst{1}, Jiarui Zhang\inst{6}, Fuxiao Liu\inst{7}}

\authorrunning{Z.~Li et al.}

\institute{~\inst{1}University of Maryland
\quad \inst{2}Brown University \quad  \inst{3}Washington University in St. Louis \\ 	\inst{4}Adobe \quad \inst{5}University of Illinois Urbana-Champaign \\ \inst{6}University of Southern California  \quad \inst{7}NVIDIA \\
\vspace{1mm} 
\email{zli12321@umd.edu \quad hongyang\_du@brown.edu \quad chengsong@wustl.edu}
}

\maketitle

\input{sections/abstract}
\label{abstract}

\input{sections/introduction}
\label{sec:intro}

\input{sections/method}

\label{sec:method}

\input{sections/experiment}

\label{sec:experiment}

\input{sections/ablation}
\label{sec:ablation}

\input{sections/relatedwork}

\label{sec:relatedwork}

\input{sections/conclusion}
\label{sec:conclusion}

%
%
\bibliographystyle{splncs04}
\bibliography{main}
\input{sections/appendix}

\label{sec:appendix}

\end{document}

%% file: sections/abstract.tex
\begin{abstract}
Self-evolving has emerged as a key paradigm for improving foundational models such as Large Language Models (LLMs) and Vision Language Models (VLMs) with minimal human intervention. While recent approaches have demonstrated that LLM agents can self-evolve from scratch with little to no data, VLMs introduce an additional visual modality that typically requires at least some seed data, such as images, to bootstrap the self-evolution process. In this work, we present Multi-model Multimodal Zero (\textbf{\ours}), the first RL-based framework to achieve zero-data self-evolution for VLM reasoning. Moving beyond prior dual-role (Proposer and Solver) setups, \ours introduces a \textbf{multi-role} self-evolving training framework comprising three specialized roles: a \textbf{Proposer} that generates abstract visual concepts and formulates questions; a \textbf{Coder} that translates these concepts into executable code (e.g., Python, SVG) to render visual images; and a \textbf{Solver} that performs multimodal reasoning over the generated visual content. All three roles are initialized from the same base model and trained using Group Relative Policy Optimization (GRPO), with carefully designed reward mechanisms that integrate execution feedback, visual verification, and difficulty balancing. Our experiments show that \ours improves VLM reasoning performance across a wide range of multimodal benchmarks. \ours establishes a scalable path toward self-evolving multi-model systems for multimodal models, extending the frontier of self-improvement beyond the conventional two-model paradigm.

  \keywords{Vision-Language Models \and Reinforcement Learning \and Self-Evolution}

  \begin{center}
\raisebox{-0.15\height}{\includegraphics[width=0.03\textwidth]{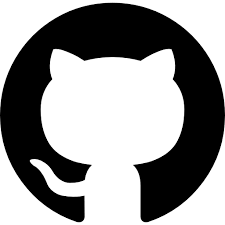}}
Code: \url{https://github.com/zli12321/MM-Zero}.
\end{center}    
\end{abstract}

%% file: sections/introduction.tex
\section{Introduction}

Self-evolving mechanisms~\citep{xiangsystematic,schmidhuber2007godel,clune2019ai,R-zero,yue2026drzeroselfevolvingsearch} represent a promising frontier for advancing foundational models.
Rather than relying on static, human-curated supervision, self-evolving systems learn by independently generating, refining, and learning from their own experiences such as through self-play~\cite{he2025visplay,wang2025visionzeroscalablevlmselfimprovement} or synthetic data generation~\citep{silver2016mastering}.
This paradigm has seen rapid progress in the domain of Large Language Models (LLMs): recent works have shown that LLMs can autonomously improve their reasoning by generating their own training tasks and refining themselves through reinforcement learning~\citep{R-zero,liu2025spice} or code execution feedback~\citep{zhao2025absolutezeroreinforcedselfplay}, achieving strong downstream performance without any human-annotated data.

A natural question is whether similar self-evolving paradigms can be extended to Vision Language Models (VLMs). 
Post-training of VLMs has traditionally relied on large volumes of expert-curated tasks and labels~\citep{liu2023visualinstructiontuning,11147946,huang2025vision, 11147946,feng2025videor1reinforcingvideoreasoning,liu2023mitigating,fei2024multimodal}, making it costly, labor-intensive, and fundamentally bottlenecked by the availability of human annotation~\citep{yang2024weak,silver2016mastering}.%
However, extending self-evolution to VLMs is non-trivial. Unlike LLMs, which operate over text alone, VLMs require visual inputs, meaning any self-evolving loop must also generate or supply images, not just questions and answers.

Existing VLM self-evolution approaches~\citep{he2025visplay,thawakar2025evolmm,wang2026v} attempt to address this by adapting the \textit{proposer--solver} pipelines originally designed for LLMs~\citep{R-zero,yue2026dr}. 
In these frameworks, a Proposer proposes visual queries and an Solver answers them. 
While this eliminates the need for human-annotated labels, the entire iterative process remains rigidly conditioned on pre-existing, collected static image datasets. 
This merely shifts the bottleneck: the model's evolution is now bounded by the distribution, quality, and diversity of the collected image dataset~\cite{he2025visplay,thawakar2025evolmm,wang2026v}. 
Self-evolving VLMs from static images thus introduces an additional burden of sourcing image data, a filtering and collection process that is itself time-consuming and costly~\citep{liu2023visualinstructiontuning,11147946,multimodalGen,Mmmu-pro}. 
In contrast, generating synthetic visual data offers a fundamentally more scalable alternative: programmatically rendered scenes can simulate far more complex and diverse scenarios than curated dataset, with virtually unlimited variations~\cite{long-etal-2024-llms,8202133}. 
This principle is already well-established in robotics, where simulated environments are widely used to train autonomous driving systems by exposing models to rare and dangerous events that would be too costly to collect in the real world~\cite{pmlr-v78-dosovitskiy17a}.



We introduce \textbf{\ours}, the first \textbf{zero-data} self-evolving reinforcement learning framework that trains VLMs without relying on any external data. Instead of conditioning on existing images, questions, or human labels, \ours enables the model to generate its own visual content, starting from simple scenes and progressively creating more complex ones.
To achieve this, we expand the traditional dual-role setup into a \textbf{tri-role} interaction, with all roles initialized from the same base model and optimized using Group Relative Policy Optimization (GRPO)~\citep{grpo}:

\begin{itemize}
    \item A \textbf{Proposer} (Abstract Conception) that formulates diverse topics, high-level visual descriptions, and associated questions;
    \item A \textbf{Coder} (Visual Synthesis) that translates these concepts into executable code (e.g., Python, SVG) to render precise visual images
    \item A \textbf{Solver} (Multimodal Reasoning) that engages with the synthesized visual content to solve complex reasoning tasks.
\end{itemize}

This tri-role architecture establishes a closed-loop system where abstract language, executable code, and visual reasoning mutually reinforce one another. Unlike previous approaches where images are fixed, the proposer is incentivized to create \textit{Goldilocks} tasks~\cite{Johnston_1990} that are challenging yet solvable for the coder to render images and the solver to solve.
our Coder receives direct \textit{execution feedback} and visual quality feedback from the solver as rewards.
Our solver uses the training data generated by the proposer and coder to perform test-time reinforcement learning (TTRL)~\cite{zuo2025ttrl} to improve itself.
To the best of our knowledge, \ours is the first framework to explore self-evolving multimodal reasoning through interactions among more than two roles, requiring zero human-annotated data.

We apply \ours to train Qwen3-VL-Instruct 4B and 8B~\cite{bai2025qwen3vltechnicalreport} as well as Mimo-VL-7B-Instruct~\cite{MiMo-VL}, demonstrating that visual reasoning capabilities can emerge purely from self-play. 
Without any real-world training data, \ours achieves consistent performance gains across diverse VLMbenchmarks. 
Our main contributions are as follows:
\begin{itemize}
\item We propose \textbf{\ours}, the first zero-data self-evolving framework that improves VLM reasoning through multi-model self-evolution, requiring no external.
\item We introduce a tri-role self-evolving pipeline (Proposer--Coder--Solver) that bridges abstract reasoning and visual grounding through intermediate code generation, demonstrating for the first time that VLMs can self-evolve with more than two roles. Our results show consistent improvements across a wide range of multimodal benchmarks and multiple base models.
\end{itemize}

%% file: sections/method.tex
\section{Methodology}
In this section, we present the reinforcement learning with verifiable rewards (RLVR) framework~\cite{wen2025reinforcement} and the self-evolving reward pipeline for \ours.

\subsection{Preliminaries}
RLVR is a training paradigm for VLMs applicable to domains where the correctness of model outputs can be verified. A rule-based verifier $v: X \rightarrow \{0,1\}$ assigns a binary reward to each generation $x_i$:
\begin{equation}
  r_i = v(x_i) =
    \begin{cases}
    1, & \text{if } x_i \text{ is correct},\\
    0, & \text{otherwise}.
    \end{cases}
\end{equation}
Such verifiable rewards are particularly effective in reasoning-intensive tasks~\cite{10.5555/3600270.3602070}, such as math and code generation~\citep{zhao2025one}, where correctness can be objectively evaluated. Beyond correctness, rewards can also encode additional criteria such as answer diversity or adherence to a desired output format.

We adopt Group Relative Pilicy Optimization (GRPO)~\citep{grpo}, a practical RL algorithm that eliminates the need for a learned value function by instead computing relative rewards across multiple samples from the same prompt. Specifically, given a prompt $p$, the current policy $\pi_{\theta_{\text{old}}}$ generates a group of $N$ responses $\{x_1, \ldots, x_N\}$ with corresponding rewards $\{r_1, \ldots, r_N\}$. These rewards are normalized within the group to yield response-level advantages:
\begin{equation}
\label{eq:advantage}
  \hat{A}_i =
\frac{r_i - \mathrm{mean}(r_1, \ldots, r_N)}
{\mathrm{std}(r_1, \ldots, r_N) + \varepsilon_{\text{norm}}},
\end{equation}
where $\varepsilon_{\text{norm}}$ is a small constant added for numerical stability. The policy is then updated by maximizing a clipped surrogate objective, regularized with a KL divergence term to constrain drift from the previous policy:
\begin{equation}
\label{eq:loss_grpo}
\resizebox{0.9\linewidth}{!}{$\displaystyle
\mathcal{L}_{\text{GRPO}}(\theta)
= -\frac{1}{N}\sum_{i=1}^{N}
\min\!\Bigg(
\frac{\pi_\theta(x_i)}{\pi_{\theta_{\text{old}}}(x_i)} \hat{A}_i,\,
\mathrm{clip}\!\left(
\frac{\pi_\theta(x_i)}{\pi_{\theta_{\text{old}}}(x_i)},
1\!-\!\epsilon,\, 1\!+\!\epsilon
\right)
\hat{A}_i
\Bigg)+ \beta\, \mathrm{KL}\!\left(\pi_\theta \,\|\, \pi_{\theta_{\text{old}}}\right)
$}
\end{equation}
By upweighting responses with positive relative advantages while penalizing excessive policy deviation, GRPO provides an effective mechanism for improving reasoning and generation quality in VLMs under the RLVR framework.

\begin{figure}[t!] 
  \centering 
  \includegraphics[width=0.99\textwidth]{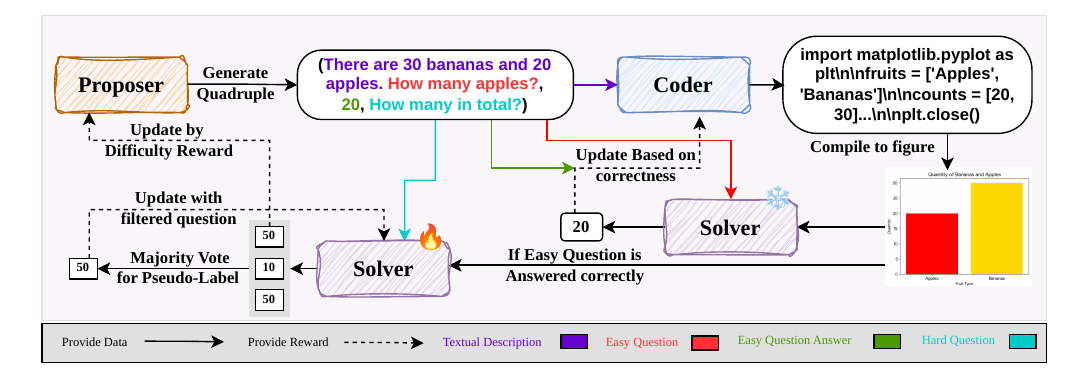}
  \caption{\textbf{Overview of the \ours self-evolving framework.} The Proposer generates a quadruple (textual description, easy question, easy answer, hard question). The Coder converts the description into executable code to render a figure. The Solver first answers the easy question to verify semantic correctness and provide reward signals to update the Coder, then answers the hard question using majority voting to generate pseudo-labels for its own training while providing a difficulty reward to optimize the Proposer.}
  \label{fig:method}
\end{figure}

\subsection{Training Pipeline}
\ours contains three distinct model agents evolved from the same base model using no additional data: a Proposer ($\pi_P$), a Coder ($\pi_D$), and Solver ($\pi_S$), as illustrated in~\cref{fig:method}. 
Each model is optimized sequentially via GRPO, with role-specific reward functions that form a closed training loop.
During each role's training phase, the other two roles \textbf{remain frozen}.
We now describe the training pipeline for each role.

\paragraph{Proposer.}
The Proposer is trained to generate visual captions and question–answer pairs. 
%
%
At each training step, the Proposer is prompted to generate captions and questions across diverse visual domains, e.g., chart understanding, object and shape recognition, OCR, and visual mathematical reasoning.
To compute rewards for the proposer, we initialize vLLM~\cite{kwon2023efficient} service ports for both the coder and solver. The coder receives the generated captions and questions and produces SVG code, which is rendered in parallel and converted to PNG images. Successfully rendered images, along with their associated easy and hard questions, are then forwarded to the solver to generate answers. The solver computes rewards for the proposer. 
We use a rollout size of 4 for the coder and 5 for the solver to reduce reward computation cost.

\paragraph{Coder.}
We use the most recent proposer checkpoint to generate approximately 4{,}000 caption and question–answer pairs, which serve as training data for the coder. The coder is trained to generate SVG code that renders images from captions. Rendered images are sent to the solver to compute reward.

\paragraph{Solver.}
Training data for the solver is constructed using the most recent proposer and coder checkpoints. The proposer generates proposals, and the coder renders them into images. We retain only successfully rendered images and their associated questions as the training set for the solver.

\paragraph{Training data filtering.} We apply stage-specific data filters to ensure training quality. For the coder, given a caption along with easy and hard questions and answers, we retain only examples where the rendering success rate across 4 rollouts falls between 0.25 and 0.75, excluding captions that are either trivially simple or too difficult to render. For the solver, we evaluate each successfully rendered image on both easy and hard questions, keeping only examples where easy-question accuracy exceeds 0.5 and hard-question accuracy falls between 0.27 and 0.75. This filtering ensures both stages are trained on examples of appropriate difficulty, neither trivially easy nor impossibly hard.

\subsection{Reward Formulations}
In this section, we describe the reward design for each of the three roles.

\subsection{Proposer Reward Formulation}
The Proposer policy $\pi_P$ generates a proposal Quadruple $x = (c, q_{\text{easy}}, a_{\text{easy}}, q_{\text{hard}})$, where $c$ is a fine-grained textual description of a visual scene, $q_{\text{easy}}$ is a simple question whose answer becomes obvious when the caption is successfully rendered into an image (used to verify rendering quality), and $q_{\text{hard}}$ is a challenging question requiring multi-step reasoning over the rendered image (used to train and evaluate the Solver).
Through training, the Proposer learns to generate richer captions and harder questions, pushing the Coder to produce more informative renderings and the Solver to perform deeper reasoning.
To evaluate $x$, we use a multi-stage and hierarchical reward mechanism that validates formatting, solvability, and difficulty via the Solver $\pi_S$. The final reward $R(x)$ is defined as:


\begin{equation}
\label{eq:proposer_reward}
R_p(x) = \resizebox{0.82\linewidth}{!}{$\displaystyle
\begin{cases}
    -1 & \text{if format invalid,} \\[6pt]
    \frac{1}{N} \sum_{i=1}^{N} \mathds{1}_{\text{exec}}(C_i)
      \cdot \bigl(\min(R_{\text{solv}}(I_i),\, 0.5) + R_{\text{diff}}(I_i)\bigr)
    + r_{\text{eh}} + r_{\text{ct}} + r_{\text{div}} & \text{otherwise.}
\end{cases}
$}
\end{equation}

Each component is defined as:

\paragraph{1. Code Generation \& Visual Rendering ($\mathds{1}_{\text{exec}}(C_i)$).}
The proposal is sent to the $\pi_D$ to produce $N$ code samples $\{C_1, \dots, C_N\}$. Each code $C_i$ is executed to render an image $I_i$. We define the execution indicator $\mathds{1}_{\text{exec}}(C_i) = 1$ if rendering succeeds, and $0$ otherwise. Failed renders contribute $0$ to the sum.

\paragraph{2. Solvability Score ($R_{\text{solv}}$).}
For a successful image $I_i$, the Solver performs $K$ rollouts on the easy question: $\{y^{(i,k)}_{\text{easy}}\}_{k=1}^K \sim \pi_S(\cdot|I_i, q_{\text{easy}})$. The solvability score is the fraction of these rollouts whose answers agree with the Proposer's intended answer $a_{\text{easy}}$, a proxy for whether the rendered image faithfully captures the caption's content:
\begin{equation}
    R_{\text{solv}}(I_i) = \frac{1}{K} \sum_{k=1}^{K} \mathds{1}\left( y^{(i,k)}_{\text{easy}} = a_{\text{easy}} \right) \in [0, 1]
\end{equation}

Each rollout's solvability is capped at $\tau_s = 0.5$ to balance between solvability and difficulty, ensuring the Proposer cannot achieve high reward through easy questions alone and must also invest in generating hard questions.

\paragraph{3. Difficulty Score ($R_{\text{diff}}$).}
Since no gold answer exists for the hard question, we adopt a test-time reinforcement learning (TTRL)~\cite{zuo2025ttrl} approach using the Solver's own answer consistency as a reward signal. The Solver performs $K$ rollouts on the hard question: $\{y^{(i,k)}_{\text{hard}}\}_{k=1}^K \sim \pi_S(\cdot|I_i, q_{\text{hard}})$. We compute the self-consistency $c_i$ as the fraction of rollouts agreeing with the majority vote $\hat{y}_i$:
\begin{equation}
    c_i = \frac{1}{K} \sum_{k=1}^{K} \mathds{1}\left( y^{(i,k)}_{\text{hard}} = \hat{y}_i \right)
\end{equation}
The difficulty reward applies the \textbf{Goldilocks} principle~\cite{Johnston_1990}:
\begin{equation}
    R_{\text{diff}}(I_i) = \min(c_i, 1 - c_i) \in [0, 0.5]
\end{equation}
This peaks at $0.5$ when the Solver is maximally uncertain ($c_i = 0.5$), indicating the question lies at the frontier of the Solver's capabilities. 

\paragraph{4. Easy-Hard Penalty ($r_{\text{eh}}$).}
To discourage the Proposer from generating hard questions that are trivially easy for the Solver, we apply a penalty when the average difficulty score across rendered images falls below a threshold $\delta_{\text{eh}}$:
\begin{equation}
    r_{\text{eh}} = \begin{cases}
        -\lambda_{\text{eh}} & \text{if } \frac{1}{|\mathcal{I}|}\sum_{I_i \in \mathcal{I}} R_{\text{diff}}(I_i) < \delta_{\text{eh}}, \\
        0 & \text{otherwise,}
    \end{cases}
\end{equation}
where $\mathcal{I}$ is the set of successfully rendered images, $\delta_{\text{eh}} = 0.15$, and $\lambda_{\text{eh}} = 0.3$.

\paragraph{5. Content-Type Diversity Penalty ($r_{\text{ct}}$).}
To encourage diversity across visual content types (e.g., charts, diagrams, math, tables), we penalize proposals whose content type is over-represented in the current batch. Let $f_t$ denote the fraction of proposals in the batch with the same content type as proposal $x$, and $\phi$ a threshold above which the type is considered over-represented:
\begin{equation}
    r_{\text{ct}} = \begin{cases}
        -\lambda_{\text{ct}} \cdot \dfrac{f_t - \phi}{1 - \phi} & \text{if } f_t > \phi, \\
        0 & \text{otherwise,}
    \end{cases}
\end{equation}
where $\phi = 0.5$ and $\lambda_{\text{ct}} = 0.15$.

\paragraph{6. Caption and Question Diversity Bonus ($r_{\text{div}}$).}
To penalize repetitive proposals within a batch, we independently cluster captions, easy questions, and hard questions using agglomerative clustering on pairwise BLEU distances. Let $s^{(\text{cap})}_x$, $s^{(\text{eq})}_x$, and $s^{(\text{hq})}_x$ denote the cluster share (fraction of the batch in the same cluster) for proposal $x$ in each field, and let $u = 1/M$ be the uniform share for a batch of $M$ valid proposals. The combined diversity adjustment is:
\begin{equation}
    r_{\text{div}} = -\text{clip}\!\Big(\big(w_c (s^{(\text{cap})}_x - u) + w_e (s^{(\text{eq})}_x - u) + w_h (s^{(\text{hq})}_x - u)\big) \cdot M \cdot \lambda_{\text{div}},\; -\lambda_{\text{div}},\; \lambda_{\text{div}}\Big)
\end{equation}
where $w_c = 0.45$, $w_e = 0.20$, $w_h = 0.35$, and $\lambda_{\text{div}} = 0.5$. Proposals in small clusters receive a bonus, while those in large clusters are penalized.

\noindent The total reward range is $[-1.0, 1.5]$. This mechanism incentivizes the Proposer to generate valid, executable scenarios where the easy question is verifiable and the hard question lies at the frontier of the Solver's capabilities, while maintaining diversity across proposals.

\subsection{Coder Reward: Execution and Validity}
The Coder $\pi_D$ receives a caption $c$ and generates code $C$. Its goal is to produce executable code that faithfully represents the Proposer's intent. The reward $R_D$ is a weighted sum of execution status, semantic correctness, and task feasibility:

\begin{equation}
    R_D(C) = R_{\text{render}} + R_{\text{solv}} + R_{\text{diff}} - \lambda_{\text{err}}
\end{equation}

where:
\begin{itemize}
    \item $R_{\text{render}} = \mathds{1}_{\text{exec}}(C) \in \{0, 1\}$ encourages valid code generation.
    \item $R_{\text{solv}} \in [0, 1]$ is identical to the Proposer's Solvability Score, ensuring the rendered image is solvable.
    \item $R_{\text{diff}} \in \{0, 1\}$ is identical to the Proposer's Difficulty Score for hard question.
    \item $\lambda_{\text{err}}$ applies penalties for syntax errors and rendering issue. -0.1 if render fails; -0.05 if code has syntax error.
\end{itemize}

\subsection{Solver Reward: Test-Time Reinforcement Learning}

The Solver policy $\pi_S$ is trained on hard questions $(I, q_{\text{hard}})$ generated by the proposer. Since ground truth labels are unavailable for these generated questions, we use Test-Time Reinforcement Learning (TTRL) via majority voting.

For a given input, the Solver generates $K$ independent reasoning paths. We identify the \textbf{silver answer} $\bar{y}$ via majority vote: $\bar{y} = \text{Mode}(\{y_1, \dots, y_K\})$. 
The reward for the $k$-th response is defined as a weighted sum of answer accuracy and structural validity:

\begin{equation}
    R_S(y_k) = \alpha \cdot R_{\text{acc}}(y_k, \bar{y}) + (1 - \alpha) \cdot R_{\text{fmt}}(y_k)
\end{equation}

where we set the accuracy weight $\alpha = 0.9$. The components are defined as:

\paragraph{1. Accuracy Reward ($R_{\text{acc}}$).}
We extract the final answer $\hat{y}_k$ from the $\texttt{\textbackslash boxed\{...\}}$ content in $y_k$ and compare it against the consensus silver answer $\bar{y}$. The reward is a binary indicator of correctness:
\begin{equation}
    R_{\text{acc}}(y_k, \bar{y}) = \mathds{1}\left( \hat{y}_k = \bar{y} \right)
\end{equation}

\paragraph{2. Format Reward ($R_{\text{fmt}}$).}
We enforce a strict structural constraint to ensure the model produces a Chain-of-Thought followed by a boxed final answer. The binary reward is:
\begin{equation}
    R_{\text{fmt}}(y_k) = \mathds{1}\left( y_k \text{ matches format } \texttt{<think>...</think> \textbackslash boxed\{...\}} \right)
\end{equation}

This mechanism incentivizes the Solver to maintain reasoning consistency while adhering to a rigorous output format.

%% file: sections/experiment.tex
\section{Experiment}
We use Qwen3-VL-4B-Instruct, 8B~\cite{yang2025qwen3technicalreport} and Mino-7B-SFT~\cite{MiMo-VL} as our base models. Training proceeds in an iterative loop over three roles (proposer, coder, and solver) saving a checkpoint for each role every 20 steps, for a total of 60 steps. All training is conducted on 8 RTX 6000/Pro 96GB GPUs.

\subsection{Solver Evaluation Data}
We evaluate the solver on a diverse range of VLM benchmarks general visual reasoning, visual mathematical reasoning, and language shortcut evaluation.

\paragraph{General Visual Reasoning.} To assess general visual understanding capabilities, we evaluate on MMMU~\cite{mmmu}, MMMU-Pro~\cite{Mmmu-pro}, ChartQA~\cite{masry-etal-2022-chartqa}, and MM-Vet~\cite{Mm-vet}. These benchmarks test general chart interpretation, object recognition, and multi-modal reasoning across diverse image types.

\paragraph{Mathematical Visual Reasoning.} Our seed topics include visual geometry, mathematical reasoning, and number problems, we evaluate on MathVerse~\cite{zhang2024mathversedoesmultimodalllm}, MathVision~\cite{bleeker2024demonstratingreducingshortcutsvisionlanguage}, MathVista~\cite{lu2024mathvista}, and VisNumBench~\cite{VisNumBench}. 
These benchmarks test the model's ability to perform rigorous quantitative and geometric reasoning over visual inputs.

\paragraph{Hallucination Detection.} Hallucination occurs when a VLM bypasses visual information and relies instead on language priors and commonsense reasoning to answer questions~\cite{Guan_2024_CVPR,li2025videohalluevaluatingmitigatingmultimodal}.
We use HallusionBench~\cite{Guan_2024_CVPR} and MMSI~\cite{yang2025mmsibenchbenchmarkmultiimagespatial}, both of which are specifically designed to test whether models ground their answers in visual content rather than defaulting to linguistic priors.

\begin{table*}[t!]
\caption{Comprehensive results on visual reasoning benchmarks. We report the results for every 20 steps throughout our iterative framework. 
The highest performance reached during training for each model is emphasized in bold. We take accuracy as the metric and Qwen2.5-14B-Instruct as a judge.}
\centering
\resizebox{\textwidth}{!}{
\begin{tabular}{@{}lcccccccccccc@{}}
\toprule
    & \multicolumn{4}{c}{\textbf{General Visual Understanding}} & \multicolumn{4}{c}{\textbf{Visual Math}} & \multicolumn{2}{c}{\textbf{Hallucination}} & \\
    \cmidrule(lr){2-5}\cmidrule(lr){6-9}\cmidrule(lr){10-11}
    \multirow{2}{*}{\textbf{Methods}} & \multirow{2}{*}{\textbf{MMMU}} & \textbf{MMMU} & \textbf{MM} & \multirow{2}{*}{\textbf{ChartQA}} & \textbf{Math} & \textbf{Math} & \multirow{2}{*}{\textbf{MathVista}} & \textbf{VisNum} & \textbf{Hallusion} & \multirow{2}{*}{\textbf{MMSI}} & \multirow{2}{*}{\textbf{Avg.}} \\
    & & \textbf{-Pro} & \textbf{-Vet} & & \textbf{Verse} & \textbf{-Vision} & & \textbf{Bench} & \textbf{Bench} & & \\
\midrule
\multicolumn{12}{@{}l}{\textit{Qwen3-VL-4B-Instruct}} \\
\quad Base Model                      & 50.2 & 41.8 & 38.5 & 79.6 & 42.3 & 33.0 & 64.0 & 45.3 & 72.3 & 26.1 & 50.2 \\
\quad \ours (Step 20, Iter 1)         & \textbf{54.8} & 44.6 & 44.5 & \textbf{83.4} & 49.8 & 35.2 & \textbf{68.5} & 49.6 & 73.2 & 26.7 & 53.5 \\
\quad \ours (Step 40, Iter 2)         & 53.7 & 45.1 & \textbf{45.4} & 83.0 & 49.4 & 36.3 & 68.3 & 49.1 & 71.5 & 25.9 & 52.8 \\
\quad \ours (Step 60, Iter 3)         & 54.4 & \textbf{45.3} & 41.7 & 83.0 & \textbf{50.0} & \textbf{37.3} & 65.7 & \textbf{49.9} & \textbf{74.2} & \textbf{28.1} & \textbf{53.4} \\
\midrule
\multicolumn{12}{@{}l}{\textit{Qwen3-VL-8B-Instruct}} \\
\quad Base Model                      & 55.8 & 46.6 & 40.8 & 76.9 & 41.6 & 31.5 & 67.7 & 47.7 & 72.8 & 25.9 & 50.7 \\
\quad \ours (Step 20, Iter 1)         & \textbf{58.7} & 49.3 & 39.0 & \textbf{78.9} & 42.8 & 36.5 & \textbf{67.7} & \textbf{55.0} & 72.1 & \textbf{29.5} & 53.0 \\
\quad \ours (Step 40, Iter 2)         & 58.2 & 51.3 & 37.6 & 78.5 & 44.2 & 38.3 & 67.1 & 54.5 & 72.3 & 29.3 & 53.1 \\
\quad \ours (Step 60, Iter 3)         & 58.3 & \textbf{53.0} & \textbf{41.7} & 79.6 & \textbf{45.1} & \textbf{39.6} & 67.2 & 53.2 & \textbf{74.1} & 28.9 & \textbf{54.1} \\
\midrule
\multicolumn{12}{@{}l}{\textit{MiMo-VL-7B-SFT}} \\
\quad Base Model                      & 57.3 & 46.1 & 39.0 & 83.7 & 46.3 & 36.6 & 70.4 & 44.7 & 55.0 & 29.6 & 50.9 \\
\quad \ours (Step 20, Iter 1)         & 56.5 & 48.4 & 43.1 & \textbf{85.5} & 55.6 & 40.9 & \textbf{73.6} & 48.0 & 70.8 & \textbf{29.9} & 55.2 \\
\quad \ours (Step 40, Iter 2)         & 59.3 & 48.7 & \textbf{48.2} & 85.0 & 56.0 & 41.5 & 73.3 & 48.3 & 70.7 & 30.6 & \textbf{56.1} \\
\quad \ours (Step 60, Iter 3)         & \textbf{60.1} & \textbf{48.7} & 45.9 & 85.0 & \textbf{56.0} & \textbf{42.5} & 72.1 & \textbf{48.5} & \textbf{71.3} & 29.9 & 56.0 \\
\bottomrule
\end{tabular}}

\label{tab:final_main_results}
\end{table*}

\subsection{Main Results}
We use LLM-as-a-judge for evaluation, as it correlates more strongly with human judgments than exact string matching. 
Specifically, we use Qwen-2.5-14B-Instruct~\cite{qwen2025qwen25technicalreport} as the judge, as its size balances judgment accuracy with inference efficiency.
We present the results of the solver in~\cref{tab:final_main_results}.

\paragraph{The solver consistently improves on diverse benchmarks over training iterations compared to the base models.}
Results in~\cref{tab:final_main_results} show that across different model sizes, average accuracy on VLM benchmarks steadily improves with additional training iterations.
Specifically, for Qwen3-VL-8B-Instruct, our solver at iteration 3 achieves an average visual math reasoning score of $54.1\%$, compared to $50.7\%$ for the base model, an improvement of 4 percentage points without any data.
While the solver also improves on general visual understanding and hallucination benchmarks, the largest gains are on visual math tasks, which demand more intensive reasoning.
These results demonstrate the effectiveness of \ours at improving VLM reasoning without any external data, pointing toward a promising direction for self-evolving multimodal models.

\begin{figure}[t!] 
  \centering 
  \includegraphics[width=0.9\textwidth]{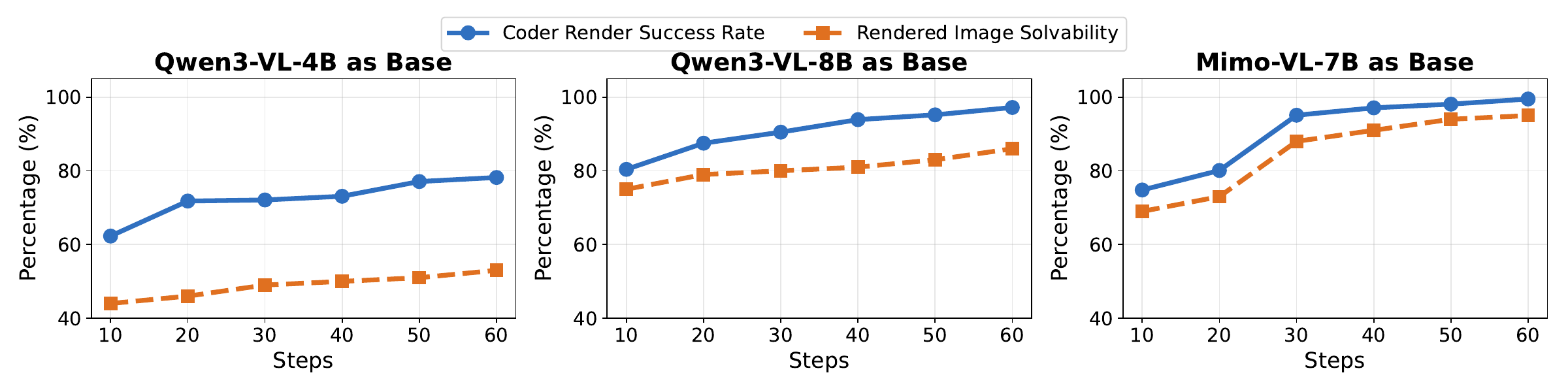}
  \caption{Throughout training, the coder's image rendering success rate steadily improves, indicating stronger code generation ability. Additionally, image solvability also increases, meaning the rendered images become more faithful to the original and contain sufficient information to answer the question.}
  \label{fig:solvability_curves}
\end{figure}

\paragraph{\ours generalizes across different models and model sizes.} As shown in~\cref{tab:final_main_results}, the average score for the 4B model increases from 50.2 to 53.4, while Mimo-VL-7B-SFT, a different model from the Qwen-VL family, improves from 50.9 to 56.0. The 4B model shows the smallest gain ($3\%$), which we attribute to its weaker base capabilities: its image rendering success rate is only around $40\%$, compared to $70\%$ for the 7B and 8B models as shwon in~\cref{fig:solvability_curves}, resulting in fewer valid training samples from the start.



%
\paragraph{Coder render success rate improves througout training.} Beyond the quantitative analysis of solver performance, we also analyze the render success rate and solvability of the generated images.
For render success rate, we track the proportion of rollouts that produce a successfully rendered image, reporting this metric every five training steps.
For solvability, we measure what percentage of successfully rendered images enable the solver to produce an answer matching the ground truth on easy questions, indicating that the rendered image contains at least some faithful visual information.
The results are shown in~\cref{fig:solvability_curves}.
Over the course of training, the coder steadily improves at generating compilable code, and the rendered images become increasingly faithful, sufficient to answer even the easiest questions correctly.

\begin{table*}[t!]
\caption{We continue training beyond three iterations on the Qwen3-VL-8B-Instruct model and observe that average benchmark scores increase monotonically through Iter 5.}
\centering
\resizebox{\textwidth}{!}{
\begin{tabular}{@{}lcccccccccccc@{}}
\toprule
    & \multicolumn{4}{c}{\textbf{General Visual Understanding}} & \multicolumn{4}{c}{\textbf{Visual Math}} & \multicolumn{2}{c}{\textbf{Hallucination}} & \\
    \cmidrule(lr){2-5}\cmidrule(lr){6-9}\cmidrule(lr){10-11}
    \multirow{2}{*}{\textbf{Methods}} & \multirow{2}{*}{\textbf{MMMU}} & \textbf{MMMU} & \textbf{MM} & \multirow{2}{*}{\textbf{ChartQA}} & \textbf{Math} & \textbf{Math} & \multirow{2}{*}{\textbf{MathVista}} & \textbf{VisNum} & \textbf{Hallusion} & \multirow{2}{*}{\textbf{MMSI}} & \multirow{2}{*}{\textbf{Avg.}} \\
    & & \textbf{-Pro} & \textbf{-Vet} & & \textbf{Verse} & \textbf{-Vision} & & \textbf{Bench} & \textbf{Bench} & & \\
\midrule
\multicolumn{12}{@{}l}{\textit{Qwen3-VL-8B-Instruct}} \\
\quad Base Model                      & 55.8 & 46.6 & 40.8 & 76.9 & 41.6 & 31.5 & 67.7 & 47.7 & 72.8 & 25.9 & 50.7 \\
\quad \ours (Step 20, Iter 1)         & 58.7 & 49.3 & 39.0 & 78.9 & 42.8 & 36.5 & 67.7 & 55.0 & 72.1 & 29.5 & 53.0 \\
\quad \ours (Step 40, Iter 2)         & 58.2 & 51.3 & 37.6 & 78.5 & 44.2 & 38.3 & 67.1 & 54.5 & 72.3 & 29.3 & 53.1 \\
\quad \ours (Step 60, Iter 3)         & 58.3 & \textbf{53.0} & 41.7 & 79.6 & 45.1 & 39.6 & 67.2 & 53.2 & 74.1 & 28.9 & 54.1 \\
\quad \ours (Step 80, Iter 4)         & 61.8 & 53.1 & 39.0 & 80.4 & 45.0 & 38.9 & 66.7 & 54.4 & 73.1 & \textbf{30.0} & 54.2 \\
\quad \ours (Step 100, Iter 5)        & \textbf{61.9} & 52.5 & \textbf{40.4} & \textbf{80.8} & \textbf{45.6} & \textbf{39.8} & \textbf{67.8} & \textbf{53.0} & \textbf{74.7} & 28.7 & \textbf{54.5} \\
\bottomrule
\end{tabular}}

\label{tab:Further Training}
\end{table*}

\paragraph{Further training of the model could lead to additional improvements.}
Beyond the initial three iterations, we continue training for two additional iterations and find that solver performance does not saturate---average accuracy continues to improve, reaching $56.6\%$ at Iter~5. This raises a natural question: \emph{how far can a model self-evolve from zero external data before reaching its ceiling without human supervision?} 
We believe this direction is central to understanding the path toward self-improving general intelligence.

\subsection{Discussion}
\begin{figure}[t!] 
  \centering 
  \includegraphics[width=1.0\textwidth]{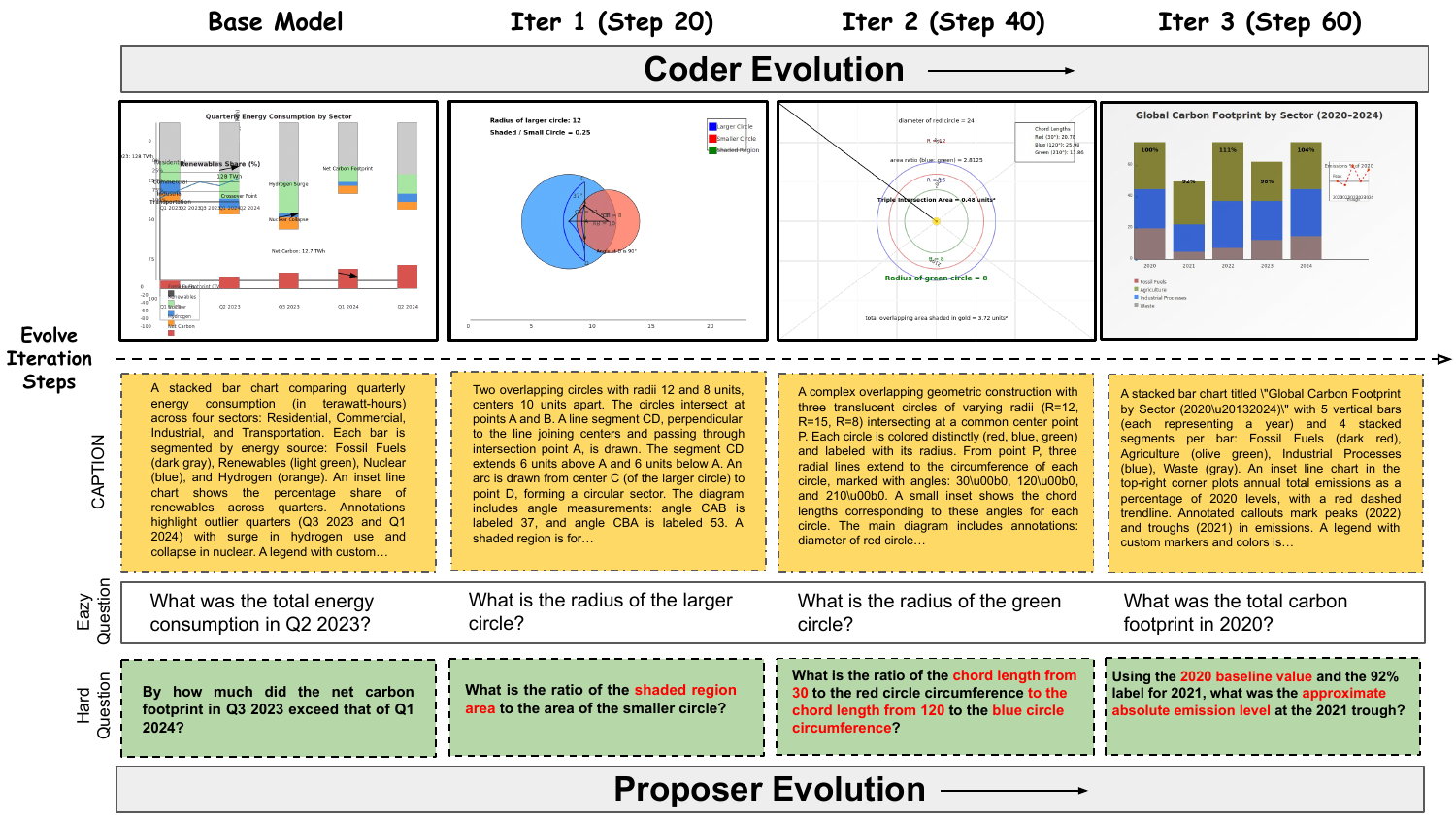}
  \caption{Evolution of the proposer and coder. This figure presents examples of questions generated by the self-evolving Vision-Language model. As the training steps progress, the visual context becomes more intricate, and the model generates increasingly challenging questions. While the upper bound of complexity grows, foundational and simple questions are still preserved. The questions are formulated by Qwen3-VL-8B-Instruct.}
  \label{fig:case_study}
\end{figure}

We present case studies for Qwen3-VL-8B in~\cref{fig:case_study}, showing proposer outputs and rendered images across training iterations. For each iteration, we sample 10 successfully rendered images and manually inspect them alongside their corresponding questions.
For the base model, generated figures suffer from cluttered layouts with overlapping elements, making them nearly unreadable and leading to low solvability rewards. By Iteration 1, visual organization improves and questions begin to demand greater reasoning, though answers are often embedded directly in the image as annotations, making them trivially easy, which is a pattern our difficulty score penalizes. By Iteration 2, visuals become cleaner and hard questions require multi-step reasoning rather than simple value extraction. By Iteration 3, layouts are more polished and questions demand genuine compositional reasoning: for instance, reading an absolute value from the y-axis, then applying a percentage annotation to compute a derived quantity.

%% file: sections/ablation.tex
\section{Ablation Study}
In this section, we conduct a reward ablation study to examine how individual reward components affect overall model performance.
Prior self-evolving frameworks, such as R-Zero~\cite{R-zero} for LLMs and Visplay~\cite{he2025visplay} for VLMs, employ only two roles and rely on difficulty and question diversity rewards. 
In contrast, \ours introduces a three-role framework that includes a coder responsible for rendering images, necessitating additional reward designs to ensure the quality of generated code and rendered images. We identify two key reward components unique to our framework and study their contributions by removing each one individually and retraining Qwen3-VL-8B-Instruct.

\paragraph{Solvability and Difficulty Balance.}
Referring to~\cref{eq:proposer_reward}, the proposer reward includes the term
$(\min(R_{\text{solv}},\allowbreak 0.5) + R_{\text{diff}})$,
where solvability is capped at~0.5.
This cap prevents the proposer from being overly rewarded for generating easy, trivially solvable instances, thereby encouraging a balance between solvability and difficulty.
In our ablation, we remove this cap and replace the term with $\bigl(R_{\text{solv}}(I_i) + R_{\text{diff}}(I_i)\bigr)$.
We train the ablated model using identical hyperparameters for three iterations.

\paragraph{Content Diversity.}
Also in~\cref{eq:proposer_reward}, the factor $r_{\text{ct}}$ measures the diversity of visual content types in the generated captions.
We include this term to incentivize the proposer to produce captions spanning a wider range of visual content categories (e.g., charts, diagrams, natural images), rather than repeatedly generating instances of a single type.

\begin{table*}[t!]
\centering
\caption{Ablation study on the solvability and difficulty balance factor and the content diversity factor. Removing solvability capping leads to slower improvement across iterations, while removing content diversity causes the solver to overfit, resulting in gradually declining accuracy.}
\label{tab:abalation_study}
\resizebox{\textwidth}{!}{
\begin{tabular}{@{}lcccccccccccc@{}}
\toprule
    & \multicolumn{4}{c}{\textbf{General Visual Understanding}} & \multicolumn{4}{c}{\textbf{Visual Math}} & \multicolumn{2}{c}{\textbf{Hallucination}} & \\
    \cmidrule(lr){2-5}\cmidrule(lr){6-9}\cmidrule(lr){10-11}
    \multirow{2}{*}{\textbf{Methods}} & \multirow{2}{*}{\textbf{MMMU}} & \textbf{MMMU} & \textbf{MM} & \multirow{2}{*}{\textbf{ChartQA}} & \textbf{Math} & \textbf{Math} & \multirow{2}{*}{\textbf{MathVista}} & \textbf{VisNum} & \textbf{Hallusion} & \multirow{2}{*}{\textbf{MMSI}} & \multirow{2}{*}{\textbf{Avg.}} \\
    & & \textbf{-Pro} & \textbf{-Vet} & & \textbf{Verse} & \textbf{-Vision} & & \textbf{Bench} & \textbf{Bench} & & \\
\midrule
\multicolumn{12}{@{}l}{\textit{Qwen3-VL-8B-Instruct}} \\
\quad Base Model                      & 53.5 & 46.9 & 39.0 & 76.6 & 41.0 & 31.0 & 67.0 & 47.3 & 73.4 & \textbf{27.5} & 50.3 \\
\midrule
\multicolumn{12}{@{}l}{\quad\textit{w/o Solvability \& Difficulty Balance}} \\
\quad\quad Step 20, Iter 1            & 55.6 & 45.5 & \textbf{43.6} & 77.4 & 41.4 & 31.6 & 66.9 & 49.7 & 71.8 & 28.0 & 51.2 \\
\quad\quad Step 40, Iter 2            & \textbf{56.2} & 45.7 & 42.2 & 76.8 & 43.4 & 35.8 & \textbf{67.7} & 54.0 & \textbf{73.0} & 28.0 & \textbf{52.3} \\
\quad\quad Step 60, Iter 3            & 54.3 & \textbf{47.1} & 41.7 & \textbf{78.1} & \textbf{43.8} & \textbf{36.5} & 67.4 & \textbf{54.4} & 71.2 & \textbf{28.1} & \textbf{52.3} \\
\midrule
\multicolumn{12}{@{}l}{\quad\textit{w/o Content Diversity}} \\
\quad\quad Step 20, Iter 1            & \textbf{57.2} & \textbf{47.7} & 40.4 & \textbf{76.9} & 42.9 & 32.6 & 66.7 & 50.6 & \textbf{73.7} & \textbf{27.9} & \textbf{51.7}\\
\quad\quad Step 40, Iter 2            & 55.8 & 46.9 & \textbf{40.8} & 73.1 & \textbf{43.4} & 34.3 & \textbf{67.1} & \textbf{51.4} & 73.2 & 26.9 & 51.3$\downarrow$ \\
\quad\quad Step 60, Iter 3            & 53.0 & 45.3 & 39.0 & 68.8 & 42.5 & \textbf{34.6} & 65.2 & 51.1 & 72.8 & 21.5 & 49.4$\downarrow$ \\
\bottomrule
\end{tabular}}
\end{table*}

We present our ablation study in~\cref{tab:abalation_study}. We analyze each condition with our perspectives.

\paragraph{Removing solvability balance leads to reward hacking by the coder.} Without the solvability and difficulty balance factor, the solver improves by only $2.3\%$ over the baseline, compared to $3.9\%$ for \ours (from $50.2\%$ to $54.1\%$). This gap stems from how the difficulty score is computed as $\bigl(\min(R_{\text{solv}}(I_i),\, 0.5) + R_{\text{diff}}(I_i)\bigr)$. When solvability is capped at $0.5$, the two components contribute roughly equally to the reward. However, without the cap, solvability can reach a maximum of $1.0$ while the difficulty optimum remains at $0.5$, creating an imbalanced reward signal. This imbalance causes the model to disproportionately optimize for solvability---generating captions that produce easy-to-solve code---while neglecting difficulty. To verify this, we randomly sample 10 generated captions and their rendered images from each iteration. We observe a consistent pattern: as training progresses, the model increasingly embeds answers directly as text within the generated code, causing the solutions to appear explicitly in the rendered images. This shortcut behavior becomes more pronounced in later iterations and is substantially mitigated when the solvability reward is capped.

\paragraph{Lack of diversity leads to overfitting to a narrow set of visual problem types.}
Without the content diversity reward, the solver improves in the first iteration ($51.7\%$) over the baseline ($50.3\%$), but accuracy steadily declines in subsequent iterations, dropping to $49.4\%$ by iteration 3.
To understand this degradation, we analyze 10 samples from iterations 2 and 3 and find that the generated captions and rendered images increasingly converge toward a narrow subset of visual types, such as histograms, which are easy to render and yield higher rendering success rates.
This pattern emerges early: without the diversity reward, the model quickly learns that certain visual types lead to higher overall rewards, and exploits this shortcut from the first iteration onward.
As a result, the model overfits to these easy-to-generate visual types, which improves reward but degrades performance on our diverse evaluation benchmarks.

%% file: sections/relatedwork.tex
\section{Related Work}
\subsection{Reinforcement Learning with Verifiable Rewards}
This paradigm has demonstrated broad success in domains where objective correctness can be programmatically evaluated, ranging from traditional mathematical and code generation tasks~\citep{guo2025deepseek,wang2025code} to complex multi-modal challenges~\citep{huang2025vision, wang2025vl,li2025self} and structured data environments~\citep{shi2025mobilegui}. To effectively leverage these deterministic feedback signals and enable complex behaviors like \ours, concurrent research is actively refining algorithmic techniques. Emphasis has shifted towards optimizing policy learning under verifiable constraints, utilizing robust frameworks like DAPO~\citep{yu2025dapo} and VAPO~\citep{yue2025vapo}, alongside high-entropy guided optimization~\citep{dai2025cde,wang2025beyond} to encourage diverse exploration and prevent premature convergence within sparse, rule-based reward landscapes.

\subsection{Self-Evolution in Vision-Language Models} The recent success of self-evolution in LLMs~\cite{xiang2026selfevolving}, with minimal or no data, has begun to extend into the VLM domain~\cite{he2025visplay,wang2025visionzeroscalablevlmselfimprovement,tong2025gamerlsynthesizingmultimodalverifiable}. 
However, unlike the single-modality setting, self-evolution in multimodal models requires generating or sourcing additional visual data for training, making the problem fundamentally more challenging~\cite{chen2023sharegpt4vimprovinglargemultimodal,xu2024llava-cot,zheng2025unicorn}. 
Existing VLM self-evolution approaches, such as VisPlay~\citep{he2025visplay}, adapt the LLM \textit{challenger-solver} pipeline to multimodal tasks but strictly require a collection of seed images to bootstrap training, which can be costly to curate and filter. 
Other concurrent works, including Evolmm~\citep{thawakar2025evolmm} and V-Zero~\citep{wang2026v}, explore multimodal self-improvement with minimal annotation but remain conditioned on pre-existing static image datasets. 
While these methods reduce the need for human-annotated labels, they remain constrained by a fixed image corpus, preventing the synthesis of progressively harder or more diverse visual scenes that could continuously challenge the model's perception and reasoning~\cite{he2025visplay,thawakar2025evolmm,wang2026v}.
Our \textbf{\ours} framework addresses this limitation by introducing a generative Coder role that programmatically renders visual content, enabling true zero-data self-evolution.

%% file: sections/conclusion.tex
\section{Limitation}
\ours demonstrates that VLMs can self-evolve by generating their own visual data and questions. Our results across multiple model sizes show that stronger base models benefit more from self-evolution, as they produce higher-quality visual code from the start—the 7B and 8B models consistently outperform the 4B model. However, the primary limitation of this work is that we were unable to validate this scaling trend on larger VLMs (e.g., 38B parameters) due to the prohibitive computational cost, leaving the full scaling behavior of self-evolving training as an important direction for future work.

\section{Conclusion}
We present \ours, a self-evolving framework that trains VLMs starting from zero data. Unlike prior self-evolving approaches, \ours instantiates more than two agents from the same base model, each fulfilling a distinct role: Proposer, Coder, and Solver. Our results demonstrate the scalability and effectiveness of multi-model multimodal evolution without any external training data: with carefully designed reward functions tailored to each role, sequential training across roles progressively improves the agents' reasoning capabilities.
Several directions remain for future work. First, the framework could be extended to support more diverse tool usage beyond code generation, enabling agents to produce a wider variety of visual components (e.g., diagrams, plots, 3D renderings) and thereby create richer and more flexible training data such as 3D spatial reasoning for the solver. 
Second, scaling to larger base models would test whether stronger initial capabilities lead to proportionally better coders and visual outputs. Finally, exploring additional agent roles beyond the current three could further enhance the self-evolving dynamics.

%% file: sections/appendix.tex
\newpage
\appendix

\begin{center}
    {\large \bf \ours: Self-Evolving Multi-Model \\ Vision Language Models From Zero Data}
    \vspace{2pt}

    {Supplementary Material}
\end{center}

\vspace{-6pt}

\section{Prompt Templates}
\label{sec:appendix-prompts}

We provide the prompt templates and training hyperparameters for reproducibility.

\subsection{Proposer}

The Proposer prompt instructs the model to act as a Visual Content
Designer and output six XML blocks:
\texttt{content\_type},
\texttt{caption},
\texttt{easy\_\allowbreak question},
\texttt{easy\_\allowbreak answer},
\texttt{hard\_\allowbreak question},
and
\texttt{hard\_\allowbreak answer}.
Allowed \texttt{content\_type} values are
\texttt{data\_\allowbreak chart},
\texttt{diagram},
\texttt{geometry},
\texttt{timeline},
\texttt{map},
\texttt{table},
and \texttt{other}.
The full prompt is shown below.

\begin{promptcard}{Proposer Prompt Template}
\textbf{Role:} You are an expert Visual Content Designer who creates
rich, complex data visualizations and diagram specifications using SVG.
Your goal is to design visualizations that are visually interesting,
data-dense, and require genuine reasoning to interpret.

\textbf{Input:} \texttt{\{\{ content | trim \}\}}

\textbf{Output --- exactly six XML blocks and nothing else:}
\begin{lstlisting}[style=promptlst]
<content_type>exactly one of: data_chart,
  diagram, geometry, timeline, map, table,
  other</content_type>
<caption>a rich, detailed specification
  </caption>
<easy_question>a simple question directly
  readable from the image</easy_question>
<easy_answer>the answer</easy_answer>
<hard_question>a challenging reasoning
  question</hard_question>
<hard_answer>the answer</hard_answer>
\end{lstlisting}
\vspace{-2pt}
\textbf{Complexity requirements} for captions --- include at least
three of:
multiple data series,
annotations,
secondary panel,
colors/markers,
derived values,
non-trivial patterns,
geometric constructions.

\textbf{Hard question constraints:}
Must require multi-step reasoning; must force the reader to extract
at least one value from the visualization (do \emph{not} state all
values in the question).

\textbf{Answer format:} single number, word, or short phrase
(e.g.\ \texttt{"42"}, \texttt{"Q1"}, \texttt{"blue"}).
\end{promptcard}

\subsection{CodeGen}

The CodeGen prompt instructs the model to generate raw SVG markup given
a caption and questions. The rendered image must contain the data needed to
answer the Easy Question. Full prompt:

\begin{promptcard}{SVG Generator Prompt Template}
\textbf{Input:} \texttt{\{\{ content | trim \}\}}

You are an SVG code generator for data visualizations.
You will be given a chart description (caption) and questions
with short answers.

\textbf{Critical:} The rendered image \textbf{must} contain the
data needed to answer the Easy Question with the exact Easy Answer
provided.

Write raw SVG markup (starting with \texttt{<svg ...>}).
Do \textbf{not} write Python code.

\textbf{SVG guidelines:}
use \texttt{viewBox};
use \texttt{<text>} for labels;
\texttt{font-size} $\geq$ 12\,px;
distinct colors;
self-contained.
Wrap your SVG in \texttt{\`{}\`{}\`{}svg ... \`{}\`{}\`{}} code fences.
\end{promptcard}

\subsection{Solver}

The Solver prompt instructs the model to think step-by-step in \texttt{<think>} tags
and give the final answer in \verb|\boxed{}|. Full prompt:

\begin{promptcard}{Solver Prompt Template}
\textbf{Input:}
\begin{lstlisting}[style=promptlst]
{% if '<image>' not in content %}
<image>{% endif %}{{ content | trim }}
\end{lstlisting}
\vspace{-2pt}
Look at the image carefully and answer the question.
First, think step by step inside
\texttt{<think>} \ldots\ \texttt{</think>} tags.
Then, give your final answer inside
\verb|\boxed{}| as a single number, single
word, or short phrase only
(e.g.\ \verb|\boxed{42}|,
\verb|\boxed{blue}|,
\verb|\boxed{Q1}|)\,---\,no units, no full
sentences.
\end{promptcard}

\subsection{LLM-as-a-Judge}

We use Qwen2.5-14B-Instruct~\cite{qwen2025qwen25technicalreport} as a judge to evaluate model outputs across all benchmarks. Specifically, we extract the final answer from each model response and prompt the judge LLM to determine whether it matches the ground-truth answer. The full evaluation prompt is shown below.

\begin{promptcard}{LLM Judge Prompt Template}
\textbf{System:} You are an answer correctness judge. Given a
question, the gold (correct) answer, and the model's answer,
determine if the model's answer is correct: equivalent to the
gold answer or semantically the same. Consider numeric equality
(e.g.\ 14 vs 14.0), option equivalence (A vs A.), and
paraphrases. Answer with exactly one word: Yes or No.

\textbf{Input:}
\begin{lstlisting}[style=promptlst]
Question: {question}

Gold answer: {gold}

Model answer: {model_answer}
\end{lstlisting}
\vspace{-2pt}
Is the model answer correct? Answer with exactly one word:
Yes or No.
\end{promptcard}

\section{Training Configuration}
\label{sec:appendix-training}

Tables~\ref{tab:main-params} and~\ref{tab:model-config} list the main script parameters and per-model training configuration. All three models use full fine-tuning (LoRA rank 0); the vision tower is trainable. Total trainable parameters: $\sim$8.77B (Qwen3-VL-8B-Instruct).

\noindent
\begin{minipage}[t]{0.48\linewidth}
\begin{promptcard}{Training Parameters}
\captionof{table}{Main script parameters.}
\label{tab:main-params}
\scriptsize
\begin{tabular}{@{}lr@{\hspace{4pt}}p{2.2cm}@{}}
\toprule
\textbf{Param.} & \textbf{Val.} & \textbf{Description} \\
\midrule
\texttt{GPU\_MEM}        & 80  & GPU mem.\ (GB) \\
\texttt{TRAIN\_STEPS}    & 10  & Steps/model/iter. \\
\texttt{NUM\_ITER}       & 3   & Self-Evolve iter. \\
\texttt{PROP\_PER\_GPU}  & 413 & Prop./GPU (P/CG) \\
\texttt{PROP\_PER\_S}    & 625 & Prop./GPU (Solv.) \\
\texttt{PROP\_ROLL\_BS}  & 18  & Prop.\ roll.\ BS \\
\texttt{GLOBAL\_BS\_P}   & 18  & Prop.\ global BS \\
\texttt{ROLLOUT\_N}      & 4   & Rolls/prop.\ (P/CG) \\
\texttt{SOLVER\_N\_R}    & 5   & Solver rollouts \\
\texttt{ROLLOUT\_BS}     & 320 & CG roll.\ BS \\
\texttt{ROLL\_BS\_S}     & 512 & Solver roll.\ BS \\
\texttt{GLOBAL\_BS\_S}   & 64  & Solver global BS \\
\bottomrule
\end{tabular}
\end{promptcard}
\end{minipage}%
\hfill
\begin{minipage}[t]{0.48\linewidth}
\begin{promptcard}{Per-Role Config (80\,GB)}
\captionof{table}{Per-role training configuration.}
\label{tab:model-config}
\scriptsize
\begin{tabular}{@{}lccc@{}}
\toprule
\textbf{Parameter} & \textbf{P} & \textbf{CG} & \textbf{S} \\
\midrule
GPUs              & 3    & 4    & 8 \\
Max prompt len.   & 4096 & 4096 & 8192 \\
Max resp.\ len.   & 2048 & 4096 & 4096 \\
Roll.\ batch size & 18   & 256  & 512 \\
Global batch size & 18   & 64   & 64 \\
Learning rate     & \multicolumn{3}{c}{$1\!\times\!10^{-6}$} \\
Weight decay      & \multicolumn{3}{c}{$1\!\times\!10^{-2}$} \\
Roll.\ temp.      & 1.0  & 0.7  & 1.0 \\
Roll.\ top\_p     & 0.99 & 0.95 & 0.99 \\
Rollout $n$       & 4    & 4    & 8 \\
Tensor parallel   & 1    & 1    & 2 \\
\bottomrule
\end{tabular}
\end{promptcard}
\end{minipage}

\section{Rendering Pipeline}
\label{sec:appendix-rendering}

The SVG-only variant converts generated SVG markup to PNG images for the Solver. Process: (1) raw SVG string $\to$ \texttt{cairosvg} $\to$ PNG; (2) per-snippet timeout 30\,s; (3) parallel workers via \texttt{ProcessPoolExecutor}. Validation: max aspect ratio 100, max dimension 16384; invalid images are discarded. Output: base64-encoded PNG for vLLM/Solver input.